\def\year{2021}\relax
\documentclass[letterpaper]{article} 
\usepackage{aaai21}  
\usepackage{times}  
\usepackage{helvet} 
\usepackage{courier}  
\usepackage[hyphens]{url}  
\usepackage{graphicx} 
\usepackage{natbib}  
\usepackage{caption} 
 \usepackage[switch]{lineno}  %

\title{Measuring the Complexity of Domains Used to Evaluate AI Systems}
\author{
    Christopher Pereyda,
    Lawrence Holder
    \\
}
\affiliations{

    Washington State University \\
    Pullman, Washington, 99163 \\
    \{christopher.pereyda, holder\}@wsu.edu 
    
}

\begin{document}

\maketitle

\begin{abstract}
There is currently a rapid increase in the number of challenge problem, benchmarking datasets and algorithmic optimization tests for evaluating AI systems. However, there does not currently exist an objective measure to determine the complexity between these newly created domains. This lack of cross-domain examination creates an obstacle to effectively research more general AI systems. We propose a theory for measuring the complexity between varied domains. This theory is then evaluated using approximations by a population of neural network based AI systems. The approximations are compared to other well known standards and show it meets intuitions of complexity. An application of this measure is then demonstrated to show its effectiveness as a tool in varied situations. The experimental results show this measure has promise as an effective tool for aiding in the evaluation of AI systems. We propose the future use of such a complexity metric for use in computing an AI system's intelligence. 
\end{abstract}

\section{Introduction}
There is currently an ever increasing number of new and innovative AI performance measures. These can range from simple tests to challenging suites of diverse problems. Yet little consideration is given to measuring and comparing the complexity of existing benchmarks and evaluation systems. This lack of cross-domain measures hinders our ability to evaluate AI systems across multiple benchmarks. Without this examination, it is difficult to assess how the field is progressing, which results in the slowing of research on general AI systems.

The AI community often focuses on the creation of an AI system with the intent of achieving better than state-of-the-art performance. This focus has resulted in spectacular AI systems that achieve a high performance on a singular task. These ideas, however, tend not to progress the AI community as a whole towards generalizable systems that are able to handle multiple diverse tasks or slightly variant domains. This lack of cross-domain examination impedes the field from achieving more generalizable results. 

To address this problem, we propose a measure of domain complexity that can be used to compare domains. Evaluating these domains with the intent of differentiation will allow us to see which AI systems are performing better when running over a suite of domains.

A measure of domain complexity is a necessary, but not sufficient, component in a framework for evaluating AI systems. In addition to a complexity measure, we also need a measure of domain similarity and a universal performance measure will be required to achieve this goal \cite{Pereyda2020MeasuringProblems}. 

The rest of the work proceeds as follows. First, we briefly examine the current state of measuring complexity in a multitude of fields. Then we propose our theory of complexity following from these works. Next, we evaluate our theory utilizing well known datasets and show how the complexity measure can be applied. Finally, we conclude by reviewing our goals and how effective the measure was to solving them.

\section{Related Work}
One approach for measuring the complexity of a certain task is to measure its Kolmogorov Complexity \cite{Wallace1999MinimumComplexity}. The Kolmogorov Complexity of a domain is the minimum size of a program which can be used to generate such a domain. This leads to the notion of compressibility. Namely, the smallest representation of the domain. While this seems like an effective way to measure complexity, it has a significant drawback. There is currently no accurate method to measure the lower bounds. Some approximations of upper bounds have been constructed \cite{Staiger1998APrediction}. One example of a problem utilizing this notion of compressibility is the Hutter Prize \cite{Melis2017OnModels}. The goal of the task is to optimize a compressibility algorithm to compress a large corpus of real world text. The creator's of the test propose that being able to compress is closely related to intelligence \cite{Hutter2004UniversalProbability}.

Another similar idea is to utilize the Minimum Message Length \cite{Wallace1999MinimumComplexity} \cite{Hernandez-Orallo2010MeasuringTest}. This is in a similar vein to the Kolmogorov Complexity. However, it does not strongly depend on a notion of maximal compressibility. These two ideas rely largely on Algorithmic Information Theory \cite{Solomonoff1964AI}. While these methods have a well understood and agreed upon theory, they remain largely impractical to use. Still some ideas can be utilized, namely minimal problem representations. That is, minimal problem representation may make for a useful measure \cite{Legg2013AnMeasure}. 

One previous attempt to measure the complexity of domains was performed by searching for a minimal network to solve the domain \cite{Pereyda2018TowardTests}. This was done by constructing a search over variable neural network architectures. The networks were then trained to find the smallest sizes that met the criterion of solved. Solving the domain was defined as achieving 95\% of the maximum possible score. While this approach is a reasonable approximation, it lacks any theoretical backing. It also heavily relies on an effective search algorithm for finding a minimal network, which may bias the results in favor of a certain domain. The method also does not allow for an effective verification of results, as each measurement is determined by a single search and not a set of searches.

Other approaches for understanding complexity come from psychology and cognitive science \cite{Hernandez-Orallo2017TheIntelligence}, where the difficulty of a domain is understood to be a function of working memory \cite{Hardman1995ProblemReasoning}. That is, how many ideas can be kept track of at once by a human. While these concepts and theories can be useful for constructing human-based difficulty measures, they do not generalize well. For instance, a computer can easily keep track of a large number of complex ideas for an indefinite period \cite{Bang2015Example-basedMemory}. This lack of generalizability does not allow for scalability or verification through the use of synthetic experimentation. 

Our approach aims to solve these keys issues. First, we create our theory of complexity. This is built off previous work and aims to guide our understanding of what it is we are measuring. This aids with understanding and validating our measure. Second, we create this measure with applicability in mind. We are able to effectively test and validate the measure and shows its practical uses. This approach helps bridge the typical problem of moving from theory to practicality. We are then able to show how our measure compares to other well-understood notions of problem difficulty.

\section{Theory of Complexity}
There is not currently a unified model to understand or describe complexity. Several theories and models have been put forward over the last few decades, but there is little consensus \cite{Hernandez-Orallo2016ComputerImplications}. Each field of study generally has its own notion of complexity, in relation to the specifics of the field. For instance, the field of psychology tends to use the amount of mental effort required to solve a problem as their complexity \cite{Logie1995Visuo-spatialMemory}. Due to this, we will focus our notion of complexity from the field of computer science.

In this field, there exists three primary perspectives for understanding complexity \cite{Liu2012TaskFramework}. The \textit{interaction} perspective defines complexity is based on the amount of interaction required to solve the task. For example, some combination of practice games and theorizing of chess to be able to solve it. The \textit{resource requirement} perspective, in which complexity is defined by the amount of total resources used in solving the task. For example, the amount of time and energy a chess player uses during a game. This is the preferred perspective for cognitive models based on working memory. The \textit{structuralist} perspective, in which complexity is defined from the set of components and rules that go into defining the task. For example, the types of pieces of a chess board and the rules that govern the actions of these pieces. One such method is to look at the size of mazes to construct a notion of difficulty \cite{Zatuchna2009LearningPerception}. This approach is often flawed by the same problems that flaw the use of Kolmogorov  Complexity \cite{legg2007universal}, in that it requires a series of approximations and assumptions to be made. 

From these differing perspectives we will focus on the resource requirement perspective. That is, we examine the policies that are used to solve a problem. This involves analysing the complexity of the policy itself to find the required resources. This is similar to measuring the memory used by an agent to solve a classification problem. This will allow us to examine how the domains and tasks are being represented by the AI system. Examining these allows us to determine how differently-capable AI systems handle the problem and to be able to weight their scores appropriately. This is not practical to do in psychology, because the human brain is not yet accurately represented at the neuronal level. Though the ideas of working memory \cite{Logie1995Visuo-spatialMemory} can be extrapolated into AI systems to better inform our measure of complexity.

To begin, we need to properly define three key spaces. The first space we will define is the \textit{task space}. This task specific space contains the intricacies of each task, including the rules that govern the task, the specific states of the task, and the resulting observable representation of the space. It is generally impossible to reduce a task to the fundamental components of a minimal representation \cite{Wallace1999MinimumComplexity}. Some exceptions exist for which the task was specifically constructed to be reducible to a minimal representation \cite{Insa-Cabrera2011EvaluatingTest}. We define the task space $\mu$ as the resulting set of possible observations.

The next space is the \textit{solution space}. The solution space is a set of resulting products from the observation space. This space holds the relevant information for defining performance measures on the AI system. For example we can examine the Cartpole solution space \cite{openaigym}. Cartpole is a reinforcement learning domain with real world physics. The goal of the domain is to balance a pole which is jointed to a cart that moves left or right for as long as possible. While each state in the space may not hold a significant amount of information, we can examine the space's depth to find the resulting score of the domain. This creates one particular performance measure, but this same space can also be used for many different measures. Using the same example, one can create a performance measure in which distance from the center is the score. That is, minimal extreme movements from the AI system increases performance. This space can be thought of as a deterministic Markov chain. The solution space is defined as a function of a particular performance metric $V$. 

The last space is the \textit{policy space}. This space governs the mapping function of the task space to the solution space. This policy space is the most interesting space as it is where AI systems live. That is, an AI system is an instantiation of a specific policy. This space holds the required information for the AI system to make selections of the task space to achieve a certain performance measure from the solution space. We will primarily focus on this space for our notion of complexity. A policy $\pi$ is defined as the mapping function from the task space to the solution space.

One theory for determining the complexity of a domain is to analyze the effort required to solve the domain \cite{Hernandez-Orallo2017TheIntelligence}. This idea has several theoretical and practical challenges that need to be addressed. Namely, the interpretation of effort required and solving the domain. While there are several cognitive theories, we will examine these ideas from an AI system perspective.

There are several methods for determining the amount of effort required to solve a domain. One such example is to use Levin's universal search to find an optimal policy within some tolerance \cite{Levin1973UniversalProblems}. This has the benefits of being well defined and accepted by the community, yet it has drawbacks. It exists as one of many search algorithms and may not be the best choice. The larger drawback is its inability to be applicable. Levin's search suffers from the same problems as Kolmogorov Complexity, in that it can only be upper approximated for most domains. This prevents us from taking full advantage of the formalisms of the search to find the effort required. 

Another approach is to examine the number of diverse instances from a task it takes for a policy to adapt to a solution \cite{Hernandez-Orallo2017TheIntelligence}. While this approach offers good insight into a variety of possible solutions, it still has flaws similar to what was discussed before. As a result we will utilize some key ideas of this to construct a more applicable complexity measure. To do this we need to further define policy.

Firstly, we need to determine the choice of policy. There exist many different paradigms of AI systems involving many different policies. Without a specific methodology for selecting a policy, the resulting solution space may be biased. Consider two types of tasks: reinforcement learning and planning. Both tasks have specific policies that can be used to efficiently solve either problem, but not the other. There exist very few policies that can effectively solve both tasks. From this, we can determine that policy choice plays an important role in determining the solution space and thus will play an important role in our notion of complexity. Therefore, we will not denote a specific policy for use in the theory, rather, we will assume an arbitrarily generalizable policy $\pi$.

Now that we have a notion of policy, we need to determine the capability of the policy. That is, the complexity of the policy space utilized in the policy. This can be thought of as the size of the resulting network of a partially observable Markov decision process. Most practical policies have a notion of capability that is intrinsic to the specific policy. For example, consider neural networks. Within each network, capability can be determine either through network topology parameters (number of layers and nodes) or as the total parameters in the network. 

This notion of policy capability is crucial to our theory of complexity since we need a practical approximation of policy size. We propose that the complexity of a domain is intrinsically tied to the smallest policy that maps the domain to a specific performance measure. That is, given a certain domain $\mu$ and performance value $V$, there exists a smallest capable policy $\pi^*$ such that, $\pi^*$ achieves $V$ on $\mu$. 

Examining only the size of the smallest policy leaves out the searching process of finding such an optimal policy. While this searching effort information may be useful to further refine the measure, we find that it is not critical in determining the measure. If a consistent methodology can be used such that the resulting search effort is constant for all policies over all domains, it can be safely ignored.

This begs the question, what performance values do we choose to examine? We postulate that there is insufficient information to make an arbitrary singular choice of performance value. As a result, we utilize the entire space of possible performance values to create a measure. We propose that Complexity is defined as the minimal representation of the policy space required to achieve a certain score over the whole interval of the solution space for a given task space.  Thus, complexity can be defined as follows:

The complexity $C$ of a domain $\mu$ is the performance measurement $V$ of a solution space $\{s\} \in \pi^*(\mu)$ resulting from a minimal policy $\pi^{*}$ operating on the domain $\mu$ for a fixed performance value $V(s)$ evaluated over all possible performance values. 

The mathematical representation for the complexity of a domain can be defined as follows:

\begin{equation}
    Complexity(\mu) = \int^{V\max_{\mu}}_{V\min_{\mu}} V^{MDL(\pi)}_{\mu}  dV
\end{equation}

Where $\mu$ is a domain, MDL is the minimal description length of a policy, $\pi$ is an agent implementing a fixed policy, $V^{\pi}_{\mu}$ is the resulting solution space score given the policy mapping, and $V$ is the score. We define $V\min_{\mu}$ and $V\max_{\mu}$ as the minimum and maximum performance value achievable over $\mu$, respectively. This depends on the particular performance metric used.

\section{Experiments}
In this section we will evaluate our theory using a variety of experiments. We compare our proposed complexity measure to other standards of information theory. The complexity measure is also applied to intuitive problems to show that the measure meets expectations. We demonstrate that domains with a relative known difficulty are effectively and accurately ranked in terms of difficulty. We also attempt to address potential biases in the measure due to population selection. Finally, a practical application of the measure is demonstrated to show its usefulness for domain analysis.

Since our measure exists only theoretically, we will make a few approximations to be able to utilize it. First is the choice of an arbitrarily generalizable policy. While there does not currently exist a truly generalizable policy, some approximations to one have been created \cite{Veness2011AApproximation}. To address this, we propose that the currently most generalizable AI system exists in the form of a neural network. Currently the standard in the field of AI research is to utilize neural networks for a diverse range of problems. Therefore, we will use a neural network as our approximation of a generalizable AI system. 

Second is how we evaluate our policies over the whole range of performance values, but this is practically infeasible. Instead, too compute the complexity measure, we create a random population of neural network based AI systems. The population varies according to each agent's capabilities. Capabilities are determined by the number of trainable parameters of the network. This is largely dependent on the number of nodes and layers which are randomly varied within each agent. Each agent in this population is given the same training period over the data. The resulting scores are compiled and used to fit a linear estimator corresponding to the agent's capabilities. This linear estimator is then used to approximate the policy curve. From this curve, we then generate an area under the curve (AuC) over the observable score ranges. This resulting AuC is then an approximation of our complexity equation. The AuC is sum-normed to 1.0 to allow for cross-domain analysis. 

All of our proceeding experiments implement the same error bars. The error is defined as a function of the resulting mean squared error (MSE) of the linear-estimator. The MSE is converted to a standard error and normalized for the number of samples. We then multiply this by 1.96 to achieve an effective standard confidence interval of 95\%. We are thus 95\% confident that the true value lies somewhere within the error bars. 

We provide an example of this methodology as follows. First we select a specific domain to evaluate its complexity. We then proceed to train a large population of neural networks with varying capabilities. The training is conducted consistently for each each agent. For example, the number of training epochs is consistent regardless of the training score achieved. This results in many capability-performance pairings. Using a linear-estimator, we approximate the curve of agent capability to performance. We then calculate the AuC of this linear-estimator for positive performance values to create a complexity measure. This measure is non-normed and must be normed to compare across domains. 

\begin{figure}[h]
    \label{fig:datasets}
    \includegraphics[width=0.5\textwidth]{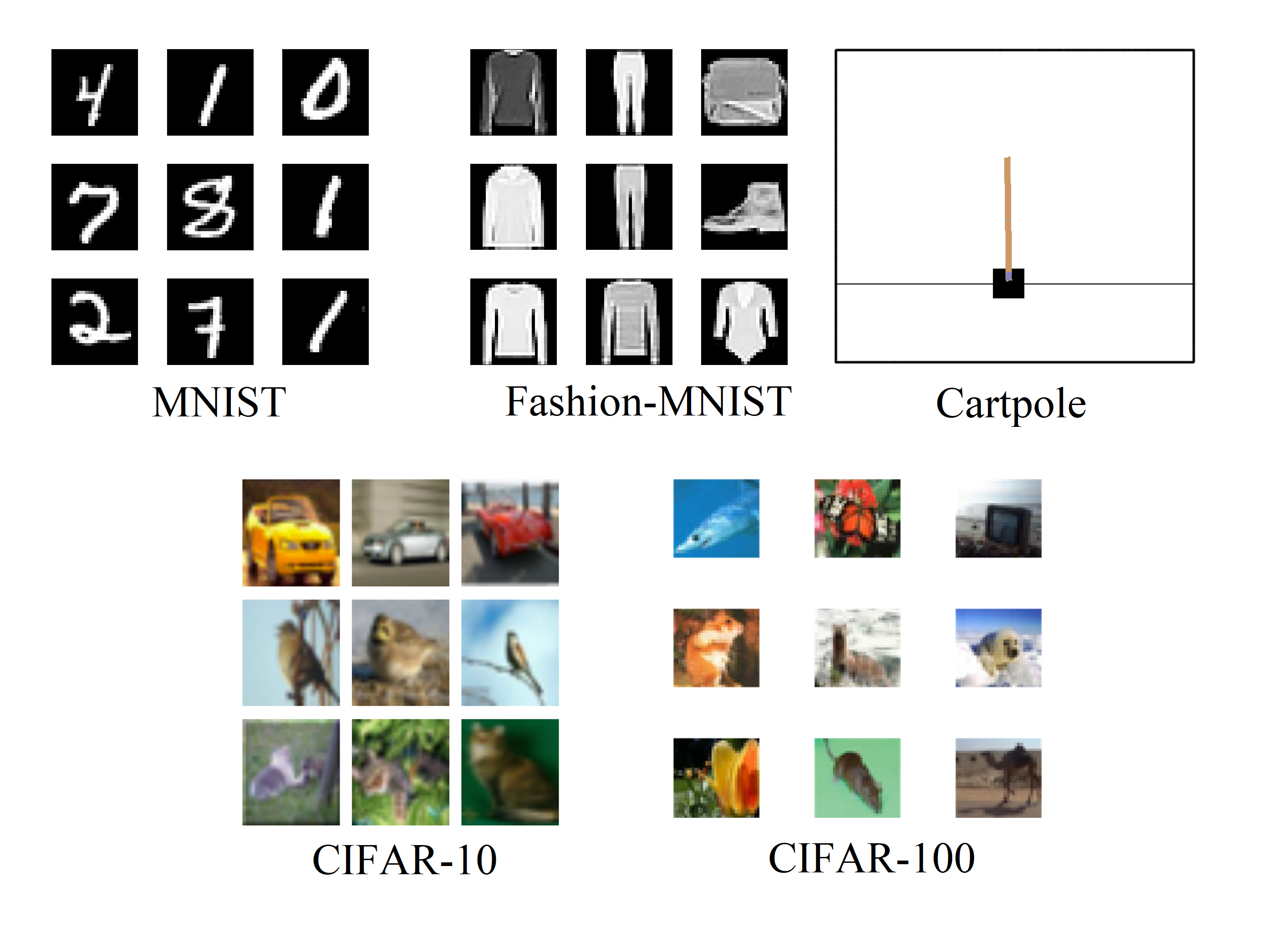}
    \caption{The five datasets used in our experimentation. Cifar-10 and Cifar-100 were converted to greyscale for uniformity. The Cartpole domain was constructed by running a trained agent on the environment to generate correct labels for the corresponding observation vector.}
\end{figure}

For these experiments we focus primarily on five well established domains. Four are image datasets: Mnist \cite{lecun1998gradient}, Fashion-Mnist \cite{xiao2017fashion}, Cifar10, and Cifar100 \cite{krizhevsky2012learning}. Each image set was converted to greyscale for ease of computation. The fifth domain is the reinforcement learning problem Cartpole \cite{openaigym}. These fice datasets can be observed in Figure 1. To fairly evaluate these domains, we transformed Cartpole into an image classification problem by running a trained agent to generate correct image-action pairings. This was done to preserve experimental consistency so we can utilize the exact same methods of measuring complexity as every other domain. While our theory is generalizable to reinforcement learning problems, we wanted to ensure the consistency of its presentation and examination within this work.

\subsection{Entropic Verification}
As mentioned before, one method for computing the difficulty of a task is to measure its Kolmorgorov Complexity. Since this complexity is not practical to measure, there are certain approximations that can be made to estimate this measure. One such method is to measure the Shannon entropy and use that as an approximation \cite{Galatolo2010EffectiveEntropy}. 

The Shannon entropy of an image is defined as the average local entropy for every pixel for a certain locality size. In our experimentation we used a radius of size 1, in most cases this leads to a locality of 5 pixels. For edge and corner pixels this size will be smaller. The entropy of this locality is calculated by determining the minimum number of bits needed to encode the locality. However, this does not account for the label portion of the dataset. To take the label into consideration, we add the base two logarithm of the number of classes in the image set, to the average entropy of the image. For example, in Cafar100 there are 100 labels to encode, so for every image in Cifar100 we add 3.32 to the average locality entropy. This final combination is the summed for every image in the set to generate a final entropy of the set.

In this experiment we measure the Shannon entropy of well known datasets and compare those to our complexity measure. For each dataset we measure the average Shannon entropy of an image. These entropies are then sum normed to one to allow for comparison to the complexity measure. These are referred to as Entropic Predictions.

We measured the complexity of each dataset using the aforementioned methods. The resulting complexity values were also sum normed to one.

\begin{figure}[h]
    \label{fig:Fig_Entropy}
    \includegraphics[width=0.5\textwidth]{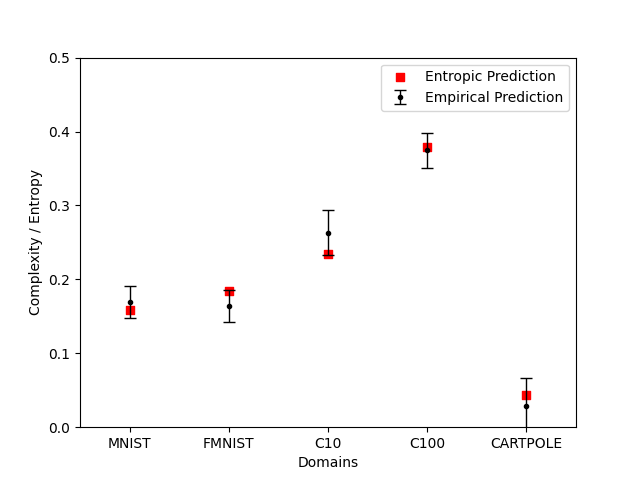}
    \caption{Entropic Prediction compared to complexity measure. We plot the average Shannon Entropy for an image in each dataset and compare that to the empirical prediction of the complexity measure. The errors bars show the 95\% confidence interval of the complexity measure.}
\end{figure}

The results are shown in Figure 2. From this data we determine that our complexity measure functions as intended. We can see that the rankings of complexity (Cifar100 being hardest and Cartpole being easiest) meets expectations. We can also determine that the complexity measurement closely approximates the entropic prediction. These approximations are within the 95\% confidence interval. Since entropy is closely tied to Kolmorgorov complexity, and the Kolmorgorov complexity of a domain is a known and useful measure of difficulty, we can conclude that our complexity measure can be utilized to measure the difficulty of a domain.

\subsection{Cluster Distribution Verification}
Another method to verify that this complexity measure captures what we expect is to show how it performs over random distributions. We can construct two normal distributions with a known mean and features corresponding to the mean and variance. The task is to predict the cluster to which a sample belongs. If we move the distribution means further apart from each other, we expect the problem difficulty to decrease. We can verify our complexity measure captures domain difficulty if the perceived difficulty decreases proportionally to the difference of the distribution means.

We construct two isotropic Gaussian clusters with a known mean and variance. We then slowly shift the means apart while applying the complexity measure. We plot the complexity measure versus the known distance. We verify the measure captures difficulty if we see a proportional increase in complexity as the means become closer. We utilize the same aforementioned methods for computing the complexity measure.

\begin{figure}[h]
    \label{Fig_Clustering}
    \includegraphics[width=0.5\textwidth]{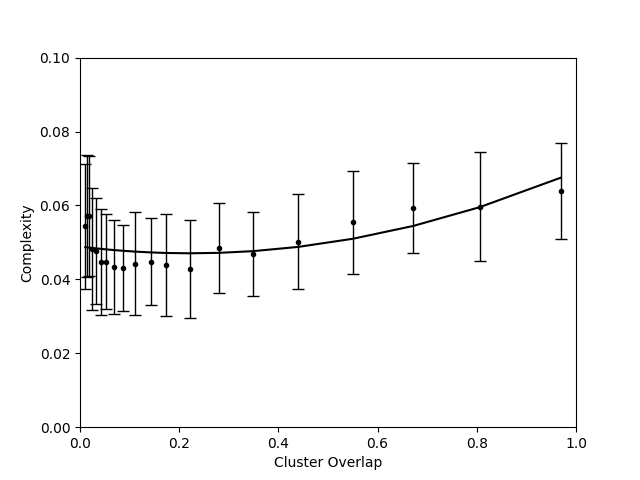}
    \caption{Measured complexity of a 2-cluster classification problem as a function of cluster overlap. The clusters are moved apart evenly using a Euclidean distance measure, however the resulting overlap between clusters is determined by the overlap in the distributions. As a result, the smaller overlapping amounts are much closer together than the larger overlaps.}
\end{figure}

Figure 3 shows the complexity versus the amount of cluster overlap. From the results we can determine that the proposed complexity measure captures difficulty. We can see that as the cluster overlap decreases, the difficulty decreases. At around 30\% overlap and lower, the complexity remains steady, because the clusters are separated enough that the features of each point in the cluster do not significantly affect the results. This is due to the isotropic nature of the cluster generating algorithm.

\subsection{Known Difficulty Verification}
A third approach to verify the proposed complexity measure is to apply it to a set of problems with a known ranking over difficulty. One such method is to take an existing dataset with multiple sets of classes and increase the number of classes. If an increase in complexity is measured alongside an increase in the number of classes then the complexity measure is validated.

This experiment was constructed utilizing the Cifar100 dataset. Random groupings of 10 classes from Cifar100 were selected and used to construct a smaller subset of Cifar. We then measured the complexity of the resulting subsets of Cifar. Since we know the possible ranges of these domains, we normalize the resulting AuC of the linear-estimator by the range of observed performance values. For example, we observe lower performance values in harder Cifar problems, so we normalize to the observed lower performance values. Since we cannot make this cross-domain assertion in other experiments, it exists only for this experiment.  

\begin{figure}[h]
    \label{Fig_Classes}
    \includegraphics[width=0.5\textwidth]{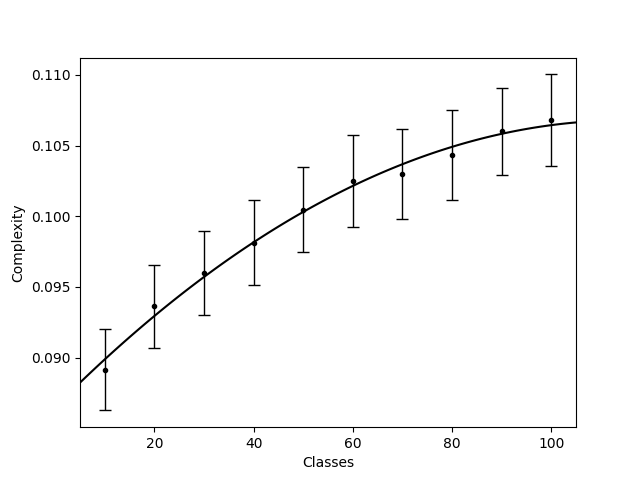}
    \caption{Complexity compared to the number of classes used in a cifar subset. All classes were selected from the original Cifar100 dataset. Each selection of classes were randomly sampled from cifar100 to prevent a certain subset of classes from biasing the results. We can observe a very clear non-linear increase in complexity as the number of classes increases.}
\end{figure}

The results of this experiment are shown in Figure 4. We can see a clearly non-linear increase of complexity as the number of classes used to construct the subset increases. We expect a non-linear curve due to the problem difficulty reaching a critical level, where adding more classes has a negligible effect on the difficulty of the problem. The results validate that the complexity measure properly ranks the Cifar problems according to known difficulty.

\subsection{Agent Dependency}
One potential problem with this experimentation is bias by not choosing from all potential agents. While it is infeasible to examine every possible type of AI system and all possible variations, it is feasible to examine how our policy search space may affect the results of our experimentation. For this experiment we utilize the five original datasets. 

We attempt to address this issue by determining how individual agents would compute the complexity measure. We then compare the similarity of these predictions to the actual complexity measure using cosine similarity. The resulting complexity fit is compared to the capability of the AI system used, as determined by the product of the number of nodes and layers. A perfect similarity between the individual AI system prediction and the actual measure is 1. 

\begin{figure}[h]
    \label{Fig_Independence}
    \includegraphics[width=0.5\textwidth]{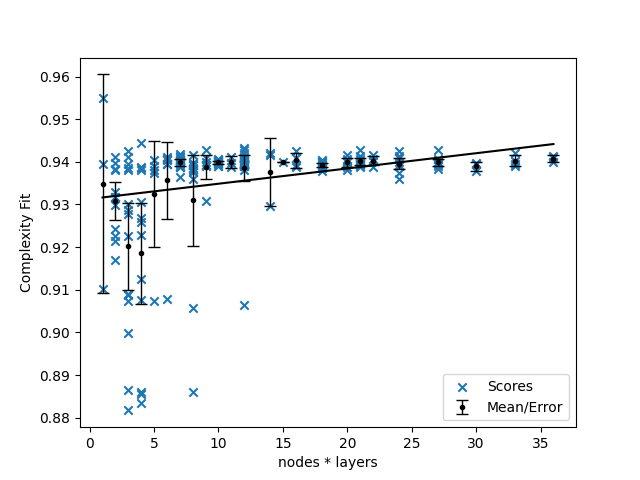}
    \caption{Complexity Fit compared to the capability of an AI system. Complexity fit is the resulting cosine similarity between the agent's score and the measured complexity over the five domains. The capability of an AI system is determined by the product of nodes times layers.}
\end{figure}

The results are shown in Figure 5. First, we observe that the resulting complexity fit is largely grouped together. That is, regardless of agent capabilities, the lowest fit achieved was 0.88. This is still very close to the target of 1. This shows that the choice of capability within an agent only matters so much to the problem. Second, we observe there is a weak positive correlation between capability and complexity fit. Measuring the Pearson correlation coefficient (PCC) yields a positive correlation of 0.36. However, we can determine that after a certain point, the variance in the results drops significantly. If we exclude capabilities below 5, we achieve a PCC of 0.16. This implies that the choice of capabilities of an AI system may have minor effects on the resulting complexity measure. If a sufficient size of capabilities are used, this effect is mitigated.

\subsection{Parameter Analysis}
Our last experiment focuses on an additional application of the complexity measure. It can be used to measure the complexity of a domain relative to another. The measure can also be used to determine how altering a certain domain parameter will affect the resulting difficulty of the domain. 

For this experiment we utilize the Cartpole domain. For each change to the domain, we alter the force applied to the cart for each action. The value is normally fixed to 10, in our experiment we vary this constant between 1.0 and 20.0 in 1.0 intervals. The resulting complexity was measured for each new force variation.

\begin{figure}[h]
    \label{Fig_force}
    \includegraphics[width=0.5\textwidth]{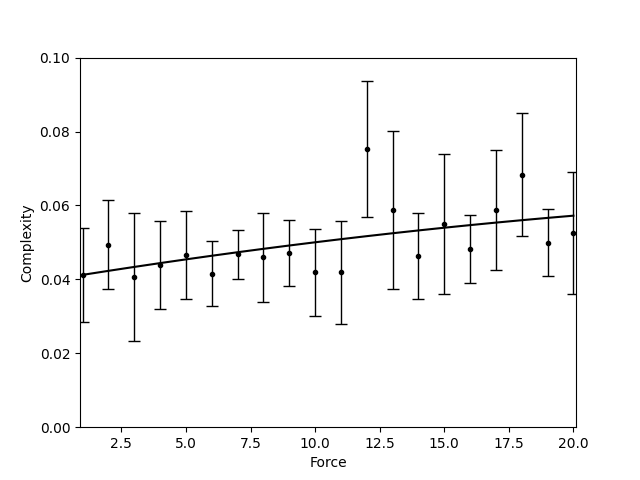}
    \caption{Complexity of the Cartpole domain compared to the change in force applied. The force applied was varied from 1.0 to 20.0 in 1.0 intervals. The complexity was measured for each new force variation. }
\end{figure}

Figure 6 shows the results of this experiment. From this data we can determine a vary clear positive correlation between force and difficulty. This makes intuitive sense as the main challenge in Cartpole is over-correction. If the force is continuously increased, the problem will get progressively harder. 

\section{Conclusion}
In this work we have constructed and put forward a novel notion of domain complexity. Within this construction we have briefly examined the theory behind it. With the goal of applicability, we have constructed an approximation of the measure that can be applied to real world domains. This approximation keeps in line with the theoretical model. 

In our experimentation we have demonstrated key aspects of this complexity measure. Through our experiments we have shown that it reasonably approximates a well known standard, Shannon entropy, which is closely related to a notion of problem complexity. We have also shown how the complexity measure meets expectations by correctly ranking problem difficulty by the number of classes involved with the problem and by correctly coorelating feature distinctiveness with difficulty. It has also been illustrated how to effectively apply this measure to examine aspect of complexity within a problem. 

While we have been able to demonstrate the complexity between domains, this was done so relatively. This methodology allows for the further research into an absolute scale of complexity. We can infer the usefulness of an absolute scale of complexity to evaluating domains in the future. Without an absolute scale, measurements made in the present may substantially drift. This will make it hard to compare domains in the future to current day domains. To solve this, we propose using this methodology on a large set of domains to construct an effective scale of complexity for all domains.

As mentioned before, this is another step to evaluating AI systems. We hold that this measure in particular will become important to a greater measure for determining AI system competence. This measure along with a notion of similarity and performance will allow for a more thorough investigation. Using a combination of three such measures will allow for the accurate evaluation of an AI system's intelligence. This can then be used to more effectively pursue generalizable AI systems. 

The AI community will greatly benefit from such a measure. Using this complexity measure, domains can be objectively determined and ordered by complexity. This will allow the further weighting of scores across a wide set of domains. From this, AI performance can be more easily evaluated to see how the field is progressing. Determining whether and by how much AI systems are progressing will make the search for generalizable AI systems significantly easier to perform.

\section{Acknowledgments}
Research was sponsored by the Defense Advanced Research Projects Agency (DARPA) and the Army Research Office (ARO) and was accomplished under Cooperative Agreement Number W911NF-20-2-0004. The views and conclusions contained in this document are those of the authors and should not be interpreted as representing the official policies, either expressed or implied, of the DARPA or ARO, or the U.S. Government. The U.S. Government is authorized to reproduce and distribute reprints for Government purposes notwithstanding any copyright notation herein.


\bibliography{mend,ref_2019}
\end{document}